\makeatletter\def\graphicscache@inhibit{true}\makeatother
\documentclass[letterpaper, 10 pt, conference]{ieeeconf}  %

\IEEEoverridecommandlockouts                              %

\usepackage{graphicx}
\usepackage[caption=false, font=footnotesize]{subfig}

\usepackage[style=ieee,hyperref,natbib=true,backend=bibtex,firstinits,doi=false,mincitenames=1,maxcitenames=2,maxbibnames=99,sorting=none,terseinits=false,hyperref=true]{biblatex}
\usepackage{bm} %
\usepackage{amsmath} %
\usepackage{amssymb}  %

\usepackage{balance}

\usepackage{float}
\usepackage{makecell}
\usepackage[normalem]{ulem}
\PassOptionsToPackage{hyphens}{url}\usepackage[hidelinks]{hyperref}
\usepackage[capitalize]{cleveref}
\usepackage{color}
\usepackage{booktabs}

\usepackage{amsfonts}
\usepackage{algorithm,algpseudocode}

\usepackage{gensymb}
\usepackage[tbtags]{mathtools}
\usepackage[pdftex,dvipsnames]{xcolor}

\usepackage{todonotes}
\usepackage{units}

\usepackage{tikz}
\usetikzlibrary{arrows}
\usetikzlibrary{positioning,calc}
\usetikzlibrary{decorations.pathreplacing}
\usetikzlibrary{decorations.markings}
\usetikzlibrary{fit}
\usetikzlibrary{shapes.callouts}
\usetikzlibrary{shapes.geometric}
\usetikzlibrary{matrix}
\usepackage[per=frac,binary-units=true]{siunitx}

\usepackage{textcomp}
\usepackage{siunitx}
\usepackage{multirow}

\IfFileExists{graphicscache.sty}{\usepackage{graphicscache}}

\usepackage{pgfplotstable}
\usepackage{pgfplots}
\pgfplotsset{compat=1.15}
\usepgfplotslibrary{groupplots}
\usepgfplotslibrary{units}
\usepgfplotslibrary{statistics}

\usepackage[draft,inline]{fixme}
\fxsetup{theme=color}

\graphicspath{{images/}}
\DeclareGraphicsExtensions{.pdf,.png,.jpg,.jpeg,.JPG}

\usepackage{dblfloatfix}

\usepackage[firstpage=true]{background}
\newcommand\copyrighttext{%
\parbox{\textwidth}{
\footnotesize
\centering
\textbf{Accepted final version.} 10th European Conference on Mobile Robots (ECMR), Bonn, Germany, September 2021
}
}
\SetBgContents{\copyrighttext}
\SetBgScale{1}
\SetBgColor{black}
\SetBgAngle{0}
\SetBgOpacity{1}
\SetBgPosition{current page.north}
\SetBgVshift{-0.8cm}
\usepackage[bottom]{footmisc}

\title{\LARGE \bf
Bimanual Telemanipulation with Force and Haptic Feedback and Predictive Limit Avoidance
}

\author{Christian Lenz* and Sven Behnke%
\thanks{\hspace{-2.2ex} *Both authors are with the Autonomous Intelligent Systems (AIS) Group, Computer Science Institute VI, University of Bonn, Germany,
        {\tt\small lenz@ais.uni-bonn.de}.
        \newline 978-1-6654-1213-1/21/\$31.00 \textcopyright 2021 IEEE}%
}%

\begin{document}
\maketitle
\thispagestyle{empty}
\pagestyle{empty}

\newcommand{\R}{\mathbb{R}}

\begin{abstract}
Robotic teleoperation is a key technology for a wide variety of applications.
It allows sending robots instead of humans in remote, possibly dangerous locations
while still using the human brain with its enormous knowledge
and creativity, especially for solving unexpected problems.
A main challenge in teleoperation consists of providing enough feedback to the human operator for situation awareness and thus create full immersion, as well as
offering the operator suitable control interfaces to achieve efficient and robust task fulfillment.
We present a bimanual telemanipulation system consisting of an anthropomorphic
avatar robot and an operator station providing force and haptic feedback to the human operator.
The avatar arms are controlled in Cartesian space with a 1:1 mapping of the operator movements.
The measured forces and torques on the avatar side are haptically displayed directly to the operator.
We developed a predictive avatar model for limit avoidance which runs on the operator side, ensuring low latency.
Only off-the-shelf components were used to build the system.
It is evaluated in lab experiments and
by untrained operators in a small user study.
\end{abstract}

\section{Introduction}

Teleoperation is a very powerful method to control robots. It enables
humans to explore remote locations and to interact there with objects and persons without being physically present.
Although state-of-the-art methods for autonomous control are improving rapidly, the experience
and instincts of humans, especially for solving unpredictable problems is unparalleled so far.
The current COVID-19 pandemic is a great example where remote work is highly desirable.
Further possible applications for teleoperation include disaster response and construction where
humans can operate remotely without risking their lives as well as
maintenance and healthcare to allow experts operating in remote locations without the need of travel.
Robotic teleoperation is a popular research area which is advanced by multiple robotic competitions like the DARPA Robotics Challenge~\citep{krotkov2017darpa}, RoboCup Rescue~\cite{kitano1999robocup} and the
ANA Avatar XPRIZE Challenge\footnote{https://www.xprize.org/prizes/avatar}.

In addition to immersive visualization of the remote location, one important aspect is telemanipulation which enables
the operator to physically interact with the remote environment. This capability is critical for many applications---without it,
we are constrained to mere telepresence.

In this work, we present a humanoid bimanual telemanipulation system built from off-the-shelf components,
which allows a human operator to interact and manipulate in remote locations. Our contributions include:
\begin{enumerate}
 \item An upper-body operator exoskeleton and a bimanual robotic avatar,
 \item an arm and hand controller with force and haptic feedback, and
 \item a model-based arm movement prediction to haptically display position and velocity limitations of the remote avatar in real time.
\end{enumerate}

\begin{figure}
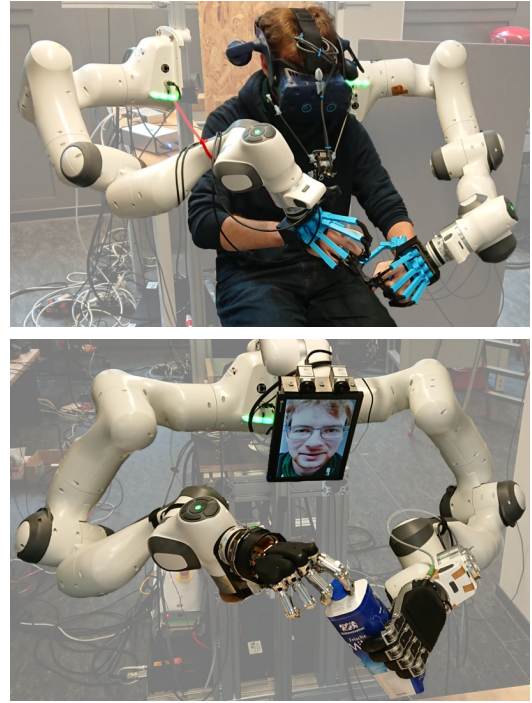

\centering
 \includegraphics[width=.8\linewidth]{images/teaser_operator.png}

 \vspace{.5em}

 \includegraphics[width=.8\linewidth]{images/teaser_avatar.png}
 \caption{Our bimanual haptic telemanipulation system: Human operator in operator station (top) performing a difficult manipulation task through a remote anthropomorphic avatar robot (bottom).}
 \label{fig:teaser}
\end{figure}

\section{Related Work}

Teleoperation is a widely investigated research area. A leading device (in our context called the Operator Station),
often with haptic feedback is used to control a following device (Avatar Robot) in a remote location.
The DARPA Robotics Challenge (DRC) 2015~\citep{krotkov2017darpa} required the development of mobile teleoperation systems.
Several research groups, such as DRC-HUBO~\citep{oh2017technical}, CHIMP~\citep{stentz2015chimp}, RoboSimian~\citep{karumanchi2017team}, and our own entry Momaro~\citep{schwarz2017nimbro} presented teleoperation systems with impressive manipulation capabilities.
The focus was on completing as many manipulation tasks as possible using a team of trained operators.
In addition, the robots were not required to communicate or interact with other humans in the remote location and thus did not feature
respective capabilities. In contrast, our developed avatar solution was designed for human interaction in the remote location and the
operator interface is designed to give intuitive control over the robot to a single, possibly untrained operator.

Some recent approaches use teleoperation interfaces which only send commands to
the robot without providing any force or haptic feedback to the operator~\citep{wu2019teleoperation,rakita2017motion}.
The advantage of such systems is clearly the light weight of the capture devices which hinder the operator only marginally.
The downside is missing force or haptic feedback, especially for tasks that cannot be solved with visual feedback alone such as difficult peg-in-hole tasks.

Other methods use custom-developed operator interfaces including force and haptic feedback~\citep{klamt2020remote,hirche2012human,guanyang2019haptic,abi2018humanoid}. In the latter,
the operator station is comparable to our approach, but the focus is placed on haptic feedback for balance control of the bipedal humanoid avatar robot.

Wearable haptic feedback devices~\citep{bimbo2017teleoperation} overcome the workspace constraints generated by stationary devices but are limited to displaying contact since they cannot create any force towards the operator.

Other research projects focus on controlling a teleoperation system under time delays \citep{niemeyer2004telemanipulation} or with two different kinematic chains on the avatar side \citep{porcini2020evaluation}.

In contrast to the highlighted related research, our approach focuses on off-the-shelf components which allow
for easy replication and maintenance. Furthermore, the used robotic arms are replaceable with any other appropriate
actuators with different kinematic chains, since the whole communication between the systems uses only the 6D end-effector pose.

\section{Hardware Setup}

\begin{figure}[]
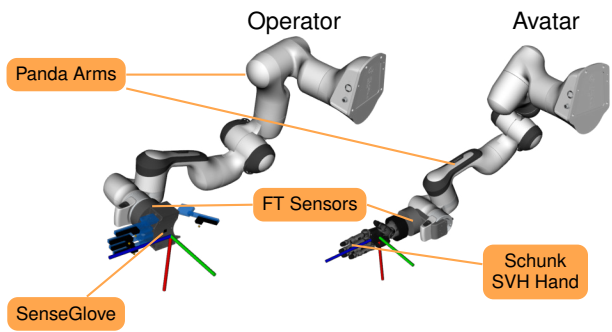

\centering
\begin{tikzpicture}[
 	font=\sffamily\scriptsize,
    every node/.append style={text depth=.2ex},
	box/.style={rectangle, inner sep=0.2, anchor=west, align=center},
	line/.style={black, thick}
]
\tikzset{every node/.append style={node distance=3.0cm}}
\tikzset{content_node/.append style={minimum size=1.5em,minimum height=3em,minimum width={width("Search Point")+0.2em},draw,align=center,fill=blue!15!white, rounded corners}}
\tikzset{header_node/.append style={minimum size=1.5em,minimum height=3em,minimum width={width("Search Point")+0.2em},align=center, rounded corners}}
\tikzset{label_node/.append style={near start}}
\tikzset{group_node/.append style={align=center,rounded corners,draw, dashed , inner sep=1em,thick}}

\node (otto)[] at (0.0,0.0) {\includegraphics[height=3.7cm]{images/otto_arm.png}};
\node (anna)[] at (3.0,0.0) {\includegraphics[height=3.3cm]{images/anna_arm.png}};

\begin{scope}[x={($ (otto.south east) - (otto.south west) $ )},y={( $ (otto.north west) - (otto.south west)$ )}, shift={(otto.south west)}]
\coordinate(otto_ft) at (0.24,0.36);
   \node[fill=orange!70,rounded corners, align=center] (senseGlove) at (0.0,0.0) {SenseGlove};
   \draw[draw=orange!70, thick] (senseGlove) -- (0.27,0.27);

   \node[fill=orange!70,rounded corners, align=center] (panda) at (0.0,0.8) {Panda Arms};
   \draw[draw=orange!70, thick] (panda) -- (0.5,0.8);

  \end{scope}

\begin{scope}[x={($ (anna.south east) - (anna.south west) $ )},y={( $ (anna.north west) - (anna.south west)$ )}, shift={(anna.south west)}]
   \node[fill=orange!70,rounded corners, align=center] (ft) at (0.0,0.35) {FT Sensors};
   \coordinate(anna_ft) at (0.32,0.29);
   \coordinate(anna_panda) at (0.45,0.5);
   \node[fill=orange!70,rounded corners, align=center] (schunk) at (0.7,0.1) {Schunk\\SVH Hand};
   \draw[draw=orange!70, thick] (schunk) -- (0.13,0.2);
  \end{scope}

  \draw[draw=orange!70, thick] (anna_ft) -- (ft.east);
   \draw[draw=orange!70, thick] (otto_ft) -- (ft.west);
   \draw[draw=orange!70, thick] (panda) -- (anna_panda);

   \node at(0.65,1.9) {\small Operator};
   \node at(4.0,1.9) {\small Avatar};
\end{tikzpicture}
 \caption{Operator (left) and avatar (right) arm with used hardware components. For simplicity, only the right arm is shown. The axes depict the common hand frame which is used for control commands and feedback.}
 \label{fig:arms}
\end{figure}

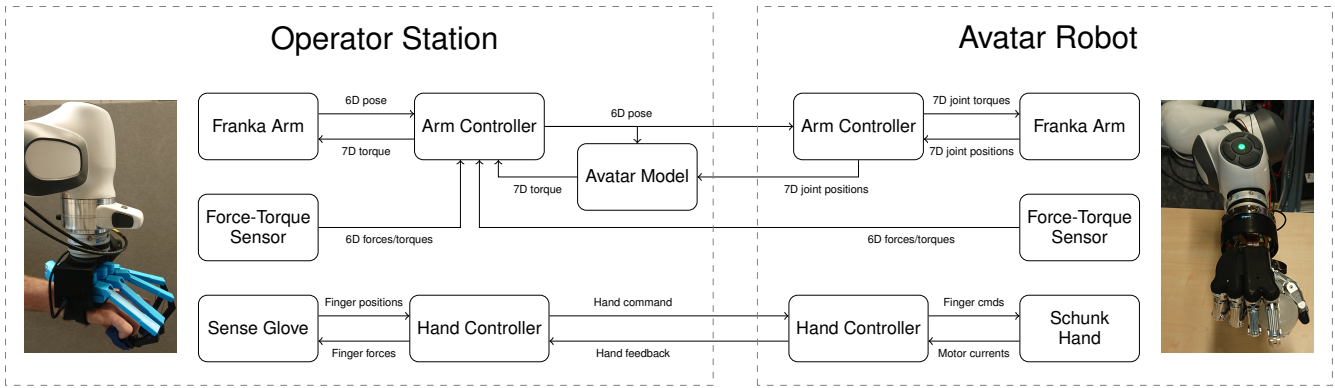
\begin{figure*}
\centering
\scalebox{0.84}{
\begin{tikzpicture}[
 	font=\sffamily\footnotesize,
    every node/.append style={text depth=.2ex},
	box/.style={rectangle, inner sep=0.2, anchor=west, align=center},
	line/.style={black, thick}
]
\tikzset{every node/.append style={node distance=3.0cm}}
\tikzset{terminal_node/.append style={minimum size=1.0em,minimum height=3em,minimum width={width("Search Point")+0.2em},draw,align=center,rounded corners}}
\tikzset{content_node/.append style={minimum size=1.5em,minimum height=3em,minimum width={width("Search Point")+0.2em},draw,align=center,fill=blue!15!white, rounded corners}}
\tikzset{header_node/.append style={minimum size=1.5em,minimum height=3em,minimum width={width("Search Point")+0.2em},align=center, rounded corners}}
\tikzset{label_node/.append style={near start}}
\tikzset{group_node/.append style={align=center,rounded corners,draw, dashed , inner sep=1em,thick}}
\tikzset{decision_node/.append style={align=center,shape aspect=1.5,minimum width=7.9em,minimum height=5.4em,diamond,draw,fill=yellow!25!white,font=\sffamily\normalsize,node distance=3.9cm}}

\node (operator)[] at (-0.5,0.0) {\includegraphics[height=4cm]{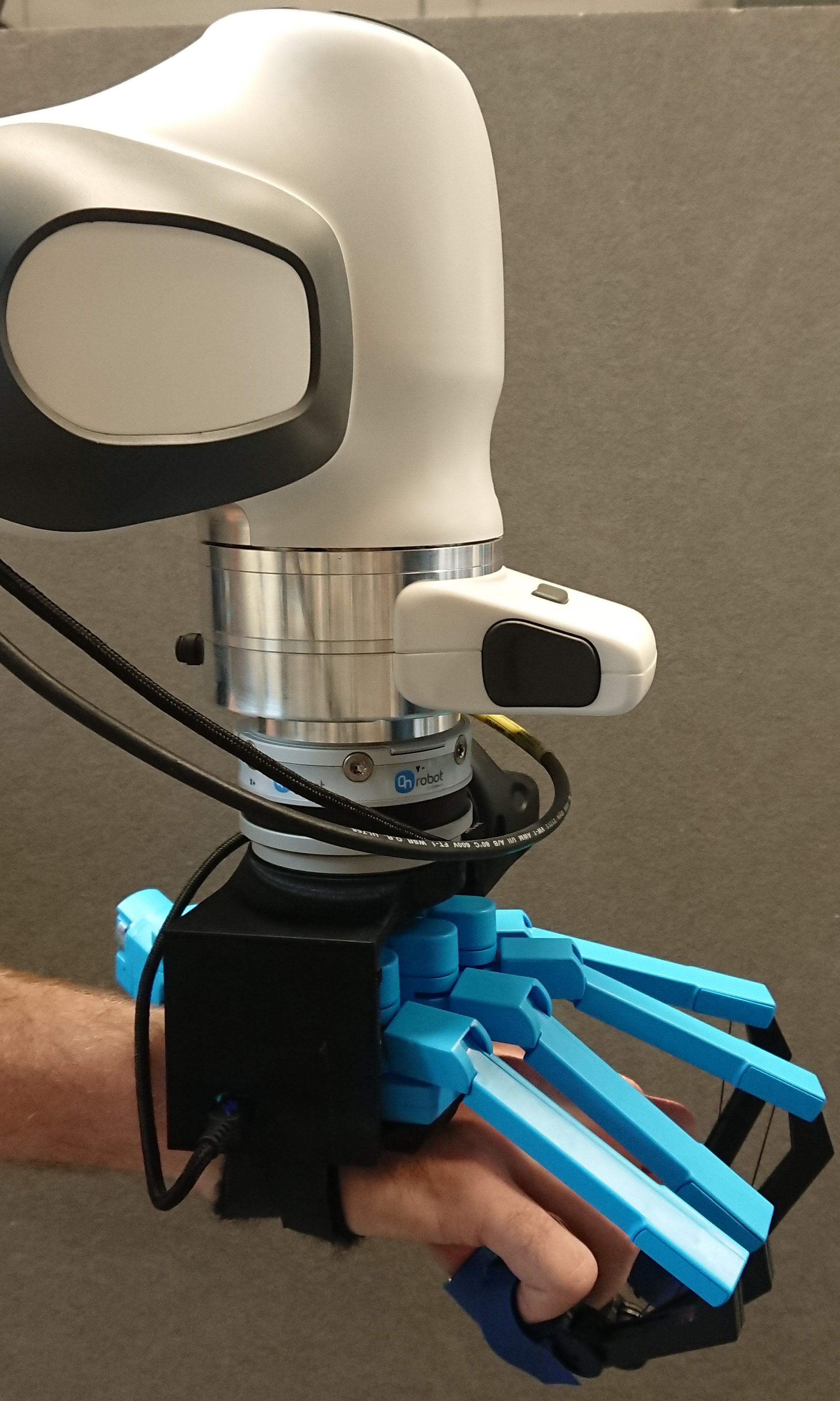}};
\node (avatar)[] at (17.5,0.0) {\includegraphics[height=4cm]{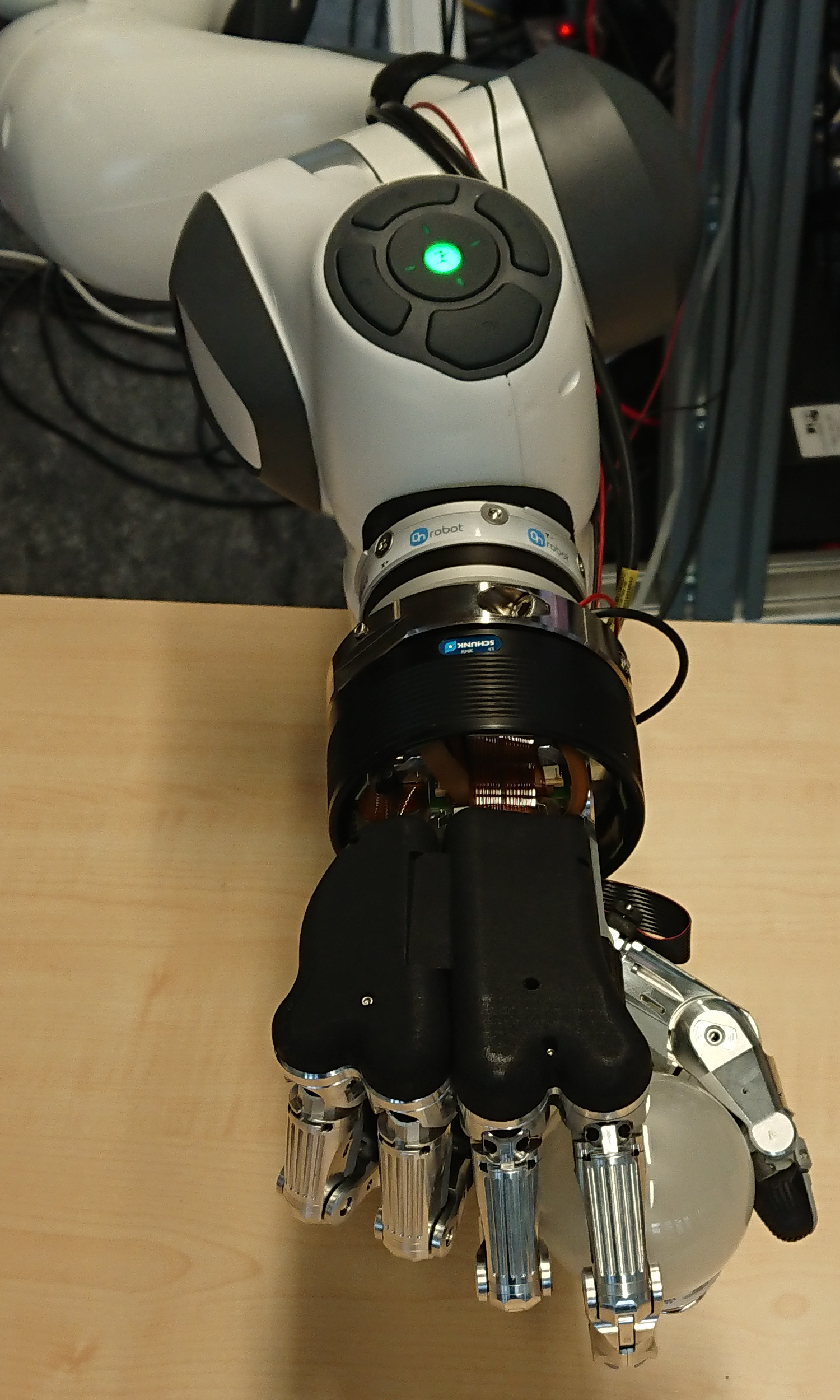}};

\node(ottoFranka)[terminal_node] at (2.0, 1.6){Franka Arm};
\node(ottoFT)[terminal_node] at(2.0,0.0) {Force-Torque\\Sensor};
\node(senseGlove)[terminal_node] at(2.0,-1.6) {Sense Glove};

\node(otto_arm_controller) [terminal_node] at(5.5, 1.6) {Arm Controller};
\node(otto_hand_controller) [terminal_node] at(5.5, -1.6) {Hand Controller};

 \draw[->]([yshift = 0.2cm] ottoFranka.east)--([yshift = 0.2cm]otto_arm_controller.west) node[midway, above]{\tiny6D pose};
 \draw[<-]([yshift = -0.2cm] ottoFranka.east)--([yshift = -0.2cm]otto_arm_controller.west) node[midway, below]{\tiny7D torque};

\node(anna_model) [terminal_node] at(8.0, 0.8) {Avatar Model};
 \draw[->](ottoFT.east) -| node[below, pos=0.25]{\tiny6D forces/torques}([xshift = -0.3cm]otto_arm_controller.south);
 \draw[->](anna_model.west) -| node[below, pos=0.25]{\tiny7D torque}([xshift = 0.3cm]otto_arm_controller.south);

 \draw [->] ([yshift = 0.2cm]senseGlove.east) -- ([yshift = 0.2cm]otto_hand_controller.west) node[midway, above]{\tiny Finger positions};
\draw [<-] ([yshift = -0.2cm]senseGlove.east) -- ([yshift = -0.2cm]otto_hand_controller.west) node[midway, below]{\tiny Finger forces};

\node(anna_arm_controller) [terminal_node] at(11.5, 1.6) {Arm Controller};
\node(anna_hand_controller) [terminal_node] at(11.5, -1.6) {Hand Controller};

\node(annaFranka) [terminal_node] at (15.0,1.6){Franka Arm};
\node(annaFT) [terminal_node] at (15.0,0.0){Force-Torque\\Sensor};
\node(schunk) [terminal_node] at (15.0,-1.6){Schunk\\Hand};

\draw [->] ([yshift = 0.2cm]anna_hand_controller.east) -- ([yshift = 0.2cm]schunk.west) node[midway, above]{\tiny Finger cmds};
\draw [<-] ([yshift = -0.2cm]anna_hand_controller.east) -- ([yshift = -0.2cm]schunk.west) node[midway, below]{\tiny Motor currents};

\draw [->] ([yshift = 0.2cm]otto_hand_controller.east) -- ([yshift = 0.2cm]anna_hand_controller.west) node[above, pos= 0.35]{\tiny Hand command};
\draw [<-] ([yshift = -0.2cm]otto_hand_controller.east) -- ([yshift = -0.2cm]anna_hand_controller.west) node[below, pos= 0.35]{\tiny Hand feedback};

\draw [->] (otto_arm_controller.east) -- (anna_arm_controller.west) node[above, pos=0.35] {\tiny 6D pose};
\draw [->] (otto_arm_controller.east) -| (anna_model.north);
\draw [<-] (anna_model.east) -| (anna_arm_controller.south) node [below, pos = 0.4] {\tiny 7D joint positions};

\draw [->] ([yshift = 0.2cm]anna_arm_controller.east) -- ([yshift=0.2cm]annaFranka.west) node [above, midway] {\tiny 7D joint torques};
\draw [<-] ([yshift = -0.2cm]anna_arm_controller.east) -- ([yshift=-0.2cm]annaFranka.west) node [below, midway] {\tiny 7D joint positions};

\draw [<-] (otto_arm_controller.south) |- (annaFT.west) node[below, pos= 0.9] {\tiny 6D forces/torques};

\draw [dashed, gray] (-2.0, -2.5) -- (-2.0, 3.5) -- (9.2, 3.5) -- (9.2, -2.5) -- (-2.0, -2.5);
\node()[header_node, align = center] at (4.0, 3.0) {\Large Operator Station};

\draw [dashed, gray] (9.9, -2.5) -- (9.9, 3.5) -- (19.0, 3.5) -- (19, -2.5) -- (9.9, -2.5);
\node()[header_node, align = center] at (14.5, 3.0) {\Large Avatar Robot};

\end{tikzpicture}
}
\caption{Control System overview. For simplicity, only the right side is depicted. The left side is controlled similarly, besides a different Hand Controller for the different Schunk hands.}
\label{fig:system}
\vspace{-1em}
\end{figure*}

The developed robotic teleoperation system consists of an operator station and an avatar robot, as shown in \cref{fig:teaser}.
The avatar robot is designed to interact with humans and in human-made indoor environments and thus
features an anthropomorphic upper body. Two 7\,DoF Franka Emika Panda arms are mounted in slightly V-shaped
angle to mimic the human arm configuration. The shoulder height of 110\,cm above
the floor allows convenient manipulation of objects on a table, as well
interaction with both sitting and standing persons. The shoulder width
of under 90\,cm enables easy navigation through standard doors.

The Panda arms have a sufficient payload of 3\,kg and a maximal reach of
855\,mm. The extra degree of freedom gives some flexibility in the
elbow position. While the arm measures joint torques
in each arm joint, we mounted additional OnRobot HEX-E 6-Axis force/torque
sensors at the wrists for more accurate force and torque measurements
close to the robotic hands, since this is the usual location of contact
with the robot's environment (see \cref{fig:arms}). The avatar is equipped with two anthropomorphic hands.
A 20\,DoF Schunk SVH hand is mounted on the right side. The nine actuated
degrees of freedom provide very dexterous manipulation capabilities.
The left arm features a 5\,DoF Schunk SIH hand for simpler but more
force-requiring manipulation tasks. Both hand types thus complement each other.

The avatar's head is equipped with two RGB cameras, a microphone,
and a small screen for communication purposes. It is attached to the
upper body using a 6\,DoF UFACTORY xArm for free head movement.
Further details on the VR remote visualization system are provided in~\citep{schwarz2021humanoid}.
The humanoid upper body has been mounted on a movable base.

The operator controls the avatar through the Operator Station from a
comfortable sitting pose. The human hand movement is captured with a similar
setup as already described for the avatar robot: The left and right Panda arms
are equipped with an OnRobot HEX-E force/torque sensor and connected to the
operator hand using a SenseGlove haptic interaction device. The Panda arm thus
serves dual purposes: A precise 6D human hand pose measurement for avatar
control, as well as the possibility to induce haptic feedback measured on the
Avatar on the human wrists. The force/torque sensor is used to measure the slightest
hand movement in order to
assist the operator in moving the arm, reducing the felt mass and friction to
a minimum.

The SenseGlove haptic interaction device features 20\,DoF finger joint
position measurements (four per finger) and a 1\,DoF haptic feedback channel
per finger (i.e. when activated the human feels resistance, which prevents
further finger closing movement).

For visual and audio communication, the operator is wearing a head mounted
display equipped with eye trackers, audio headset and a camera viewing the
lower face part (for more details see~\citep{avatar2021iros}). The avatar locomotion will be controlled using a 3D Rudder
foot paddle device.

The Panda arms feature built-in safety measures and will stop immediately
if force, torque, or velocity limits are exceeded. This ensures safe human-robot
interactions both on the operator and the avatar side.

The operator station and the avatar robot are controlled with a standard desktop
computer (Intel i9-9900K @ 3.60\,GHz, NVidia RTX 2080) each.
The communication between these computers is achieved by a single Gigabit
Ethernet connection.

\section{Force Feedback Controller}

The control architecture for the force feedback teleoperation system consists
of two arm and hand controllers (one for each the operator and the avatar side). For
the right and the left arm each controller pair is running separately. The hand controller
for the right and left hand are slightly different since different robotic hands are used.
An overview of the control architecture is shown in \cref{fig:system}.
The arm controllers run with an update rate of 1\,kHz and the force-torque sensor
measurements are captured with 500\,Hz. The force-torque measurements are smoothed
using a sensor-sided low-pass filter with a cut-of frequency of 15\,Hz.

Since the robot arms attach from outside to the operator's wrists (see \cref{fig:system}),
the kinematic chains of avatar and operator station differ,
and thus a joint-by-joint mapping of the operator
and avatar arm is not possible. Consequently, the developed control concept does not rely on similar
kinematic chains. Instead, a common control frame is defined in the middle of
the palm of both the human and robotic hands, i.e. all necessary command and
feedback data are transformed such that they refer to this frame before being
transmitted.
The controllers for the operator and avatar arm and both hands are described
in the following.

\subsection{Operator Arm Control \& Predictive Limit Avoidance}
\label{sec:operator_arm_control}
The operator arm controller commands joint torques to the Panda arms and
reads the current operator hand pose to generate the commanded hand pose sent to
the avatar robot. The goal is to generate a weightless feeling for the
operator while moving the arm---as long as no haptic feedback is displayed.
Even though the Panda arm has a quite convenient teach-mode using the internal
gravity compensation when zero torques are commanded, the weightless feeling
can be further improved by using precise external force-torque measurements.
For simplicity, just one arm is mentioned in the following, since the left
and right arms are controlled equally.

Let us denote with $\tau_o \in \R^7$ the commanded joint torques for a particular time step. Then
\begin{equation}
 \tau_o = \alpha \tau_{cmd} + \tau_{h} + \tau_{lo} + \tau_{la} + \tau_{no} + \tau_{co}\label{eq:operator_control}
\end{equation}
describes the used torque components (command, haptic feedback, operator limit avoidance, avatar limit avoidance, null space, and Coriolis) which will be explained in the following.
Note that the gravity compensation is not taken into account here, since it is
done by the Franka Control Interface (FCI) itself.

The commanded joint torques $\tau_{cmd}$ to move the Panda arm based on the
force/torque sensor measurements are defined as

\begin{equation}
\tau_{cmd} = J^T F
\label{m:tau_cmd}
 \end{equation}
with $J$ being the body Jacobian relative to the hand frame and $F\in\R^6$ the
measured 3D forces and 3D torques.
Note that $F$ has to be corrected taking sensor bias and attached end-effector
weight into account, as well as transformed into the common hand control frame.

$\tau_h$ denotes the force feedback induced by the avatar-side force-torque
sensor. Since the measurements are already bias-corrected and correctly transformed,
\cref{m:tau_cmd} can be directly applied analogously to compute
induced joint torques.

\begin{figure}[t]
  \centering
  \begin{tikzpicture}[font=\footnotesize\sffamily]
  \centering
  \begin{axis}[height=4cm,width=.9\linewidth,
     xlabel={$d_p$ [rad]}, ylabel={$\alpha$},
     xmin=0,xmax=5,ymin=-0.1,ymax=1.1,
     xticklabels={,,$t_p$,,$\frac{1}{2} t_p$,,0},
     xtick pos=left,
     ]
     \draw [black] (0.0, 1.0) -- (1.0, 1.0);
     \draw [black] (1.0, 1.0) -- (3, 0.0);
	 \draw [black] (3.0, 0.0) -- (5, 0.0);
   \end{axis}
  \end{tikzpicture}
  \vspace{-1em}
  \caption{Operator arm torque command (see \cref{eq:operator_control}) is scaled using $\alpha$ to reduce oscillations when getting close to joint position and velocity limits. The scalar decreases linearly if the distance to a joint position limit $d_p$ exceeds the threshold $t_p$. Velocity limits are handled analogously.}
  \label{fig:alpha}
\end{figure}
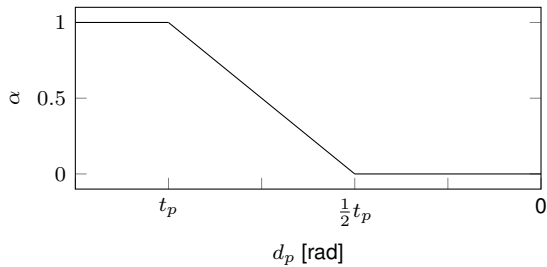

Humans can achieve high speeds moving their arm which can exceed the Panda joint velocity limits (up to $150^\circ / sec$).
In order to prevent the operator from exceeding joint position or velocity limits
of the Panda arm, the term $\tau_{lo}\in\R^7$ is introduced to apply torques
pushing the arm away from those limits. For a single joint $i$ the torque to
avoid its position limit is defined as

\begin{alignat}{2}
 \tau^i_{lo-position} = \left\{\begin{array}{ll} \beta_p (\frac{1}{d^i_p} - \frac{1}{t_p}), & d^i_p < t_p \\
         0, & \text{else}\end{array}\right.
\end{alignat}
with $\beta_p$ being a constant scalar, $d^i_p$ is the distance for
joint $i$ to its closer position limit, and $t_p = 10^\circ$ is a threshold
how close a joint has to be at a limit to activate this behavior. $\tau_{lo-velocity}^i$
is calculated analogously with $t_v = 40^\circ / sec$. Together, $\tau_{lo}$ is defined as
\begin{equation}
 \tau_{lo} = \tau_{lo-position} + \tau_{lo-velocity}. \label{eq:avoidance}
\end{equation}
The torques $\tau_{lo}$ exhibit hyperbolical growth when getting closer to respective limits.
Since the operator-side force-torque sensor will measure the generated limit
avoidance torques, the arm can end up oscillating, especially being close to one
or multiple position limits. Thus, the torques $\tau_{cmd}$, which are
influenced by the force-torque sensor, are scaled per joint by $\alpha$, which is defined as

\begin{equation}
\alpha = \max(0, \min(1, 2 \min(\frac{d_p}{t_p}, \frac{d_v}{t_v}) - 1)).
\end{equation}
The scalar $\alpha$ is designed to decrease linearly and reach zero when the limit is approached half way after activating the limit avoidance (see \cref{fig:alpha}). This reduces the commanded torques $\tau_{cmd}$ enough when approaching a position or velocity limit and prevents the oscillation.

As already mentioned, the operator station and avatar robot have different kinematic arm chains. Therefore, avoiding position and velocity limits on the operator side does not guarantee limit avoidance on the avatar side. Calculating the joint torques preventing joint limits on the avatar robot in a similar way is not beneficial, since the feedback information would arrive with high latency.
To overcome this problem, we use a model of the avatar inside the operator arm controller to predict the avatar arm movement for the next time cycle and calculate the needed joint torques to prevent joint limit violations in advance.
The model is updated with the current measured avatar joint positions each time cycle. After calculating the desired 6D avatar arm pose, the model is used to approximate the next joint positions using inverse kinematics. The current joint velocities are approximated by a low-pass filtered version of the joint position first derivative.
Having estimated the joint positions and velocities, we can apply the same avoidance strategy as described above (see \cref{eq:avoidance}).
Finally, the resulting joint torques can be transformed into the common 6D hand frame using the pseudoinverse of the Jacobian transpose $(J_A^T)^+$ and back to joint torques for the operator arm with $J_O^T$. This results in
\begin{equation}
 \tau_{la} = J_O^T (J_A^T)^+ \tau_{la-model}.
\end{equation}

The torque component $\tau_{no}$ is a null space optimization term which pulls the elbow
towards a defined convenient pose in the null space of the Jacobian. The result is a more human-like
elbow pose to maximize the operator workspace by pushing the arm away from singularities.
The last torque component is the Coriolis term $\tau_{co}$ obtained by the Panda model.

\subsection{Avatar Arm Control}
\label{sec:avatar_arm_control}

The avatar arm controller commands the Panda arm on the avatar robot to follow the commanded 6D
pose by sending joint torques $\tau_a\in\R^7$ to the Franka Control Interface. The commanded torque is defined as
\begin{equation}
 \tau_a = \tau_{cmd} + \tau_{init} + \tau_{na} + \tau_{ca}.
\end{equation}

The components $\tau_{na}$ and $\tau_{ca}$ are the null space optimization and Coriolis terms similar to $\tau_{no}$ and $\tau_{co}$ as described in \cref{sec:operator_arm_control}.
The convenient elbow poses used for the nullspace optimization is defined such that
the elbows are slightly stretched out. This generates a more human-like arm configuration
and keeps the arm away from singularities.

The goal torques $\tau_{cmd}\in\R^7$ and $\tau_{init}\in\R^7$ are generated using a Cartesian impedance controller that emulates a spring-damper system. Its equilibrium point is the 6D goal pose commanded by the operator station in the common hand frame:
\begin{equation}
 \tau_{cmd} = J^T (-S \Delta p - D (J \dot{q})),
\end{equation}
where $J$ denotes the zero Jacobian, $S\in\R^{6\times6}$ and $D\in\R^{6\times6}$
denote the stiffness and damping matrix, $\Delta p \in\R^6$ is the error in translation
and rotation between the current and goal end-effector pose and $\dot q\in\R^7$ denotes
the current joint velocities.
$\tau_{init}$ is only used for a safe initialization procedure (see below) and generated similar using the current end-effector pose.

The stiffness and damping parameters are empirically tuned
to achieve some compliance while still closely following the operator command.
When no goal pose command is received (after a communication breakdown or after the initial start),
the controller keeps commanding the current arm pose to remain in a safe state. After
receiving a command, the controller performs an initialization procedure which
fades linearly between the current and new received goal pose. This prevents the robot
from generating high torques in order to suddenly reach the new, possibly distant pose.
This initialization process takes about 3\,sec.

The Panda arm stops immediately when excessive forces are measured, for example when
there is unintended contact that exceeds force/torque thresholds. This feature is necessary to operate in a safe way.
After notification of the human operator, the avatar arm controller can restart
the arm automatically.
After performing the initialization procedure, normal teleoperation can be resumed.

\subsection{Hand Control}

The operator finger movements are captured using two SenseGlove haptic
interaction devices. Four separate finger
joint measurements are provided per finger. Since the Schunk robotic hands
on the avatar have nine and five
actuated joints, respectively, only the corresponding joint measurements are selected and
linearly mapped to the
avatar hands. While this mapping does not precisely replicate hand
postures---this is impossible anyways due to the
different kinematic structure---it gives the operator full control over all hand DoFs.

Both hands provide feedback in the form of motor currents, which is used
to provide per-finger haptic feedback to the operator.
The SenseGlove brake system is switched on or off depending on a pre-defined
current threshold.

\subsection{Force-Torque Sensor Calibration}

\begin{figure}
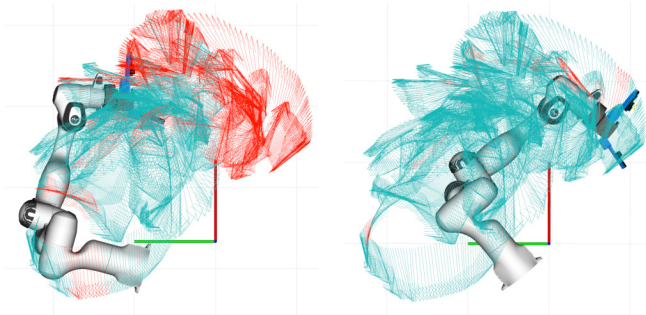

 \includegraphics[width=.48\linewidth]{images/experiments/now_final.png}\hfill
 \includegraphics[width=.48\linewidth]{images/experiments/new_final.png}

 \caption{Arm workspace evaluation. Left: Initial arm setup, similar to the avatar side. Right: Optimized mounting pose. Turquoise (reachable) and red (not reachable) arrows depict the captured human left-hand poses. The coordinate axes depict the operator sitting pose.}
 \label{fig:exp_arm_reach}
\end{figure}

\pgfplotstableread{exp/feedback1.txt}\feedbacktable
\pgfplotstableread{exp/model1.txt}\modeltable
\begin{figure}[b]
  \centering
  \begin{tikzpicture}[font=\footnotesize\sffamily]
   \begin{axis}[height=4cm,width=.9\linewidth,
     xlabel={Time [s]}, ylabel={$q_1$ [rad]},]
     \addplot+ [mark=none, green, each nth point=10, filter discard warning=false, unbounded coords=discard] table [x=Time,y=j1] {\feedbacktable};
     \addplot+ [mark=none, blue, each nth point=10, filter discard warning=false, unbounded coords=discard] table [x=Time,y=j1] {\modeltable};
     \draw [black] (1.4, -1.08) -- ( 1.4, -0.98);
     \draw [black] (1.6, -1.08) -- ( 1.6, -0.98);
     \draw[latex-latex] (1.4,-1.0) node [anchor=east] {$\Delta t$} -- (1.6,-1.0) ;
   \end{axis}
  \end{tikzpicture}
  \vspace{-1em}
  \caption{Predictive avatar model: Measured joint position for the first joint of the right avatar arm during a grasping motion (green) and predicted
  joint position for predictive limit avoidance (blue). Both measurements are captured on the operator side. Communication between both systems and motion execution generate a delay of up to 200\,ms ($\Delta t$) which
  is compensated by the predictive model.}
  \label{fig:exp_model}
\end{figure}
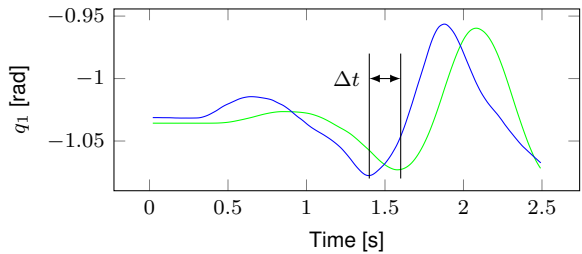

\pgfplotstableread{exp/ft2.txt}\fttable
\pgfplotstableread{exp/franka2.txt}\frankatable
\begin{figure*}[h]
  \centering
  \begin{tikzpicture}[font=\footnotesize\sffamily]
   \begin{axis}[height=4cm,width=.5\linewidth,
     xlabel={Time [s]}, ylabel={Force [N]},
     ymin = -15, ymax =15,]
     \addplot+ [mark=none] table [x expr=\coordindex,y=forceZ] {\fttable};
     \draw (30.0, 10.0) node [anchor=east, blue] {Ours};
   \end{axis}
  \end{tikzpicture}
\begin{tikzpicture}[font=\footnotesize\sffamily]
   \begin{axis}[height=4cm,width=.5\linewidth,
     xlabel={Time [s]},
     ymin = -15, ymax =15,]
     \addplot+ [mark=none] table [x expr=\coordindex,y=forceZ] {\frankatable};
     \draw (30.0, 10.0) node [anchor=east, blue] {Panda};
   \end{axis}
  \end{tikzpicture}
  \vspace{-1em}
  \caption{Operator arm movement: Force in z-direction (in the direction of the human palm) needed to move the arm in the same repetitive motion with our operator arm controller running (left)
  and while using only the Panda built-in gravity compensation (right).}
  \label{fig:exp_ft}
\end{figure*}
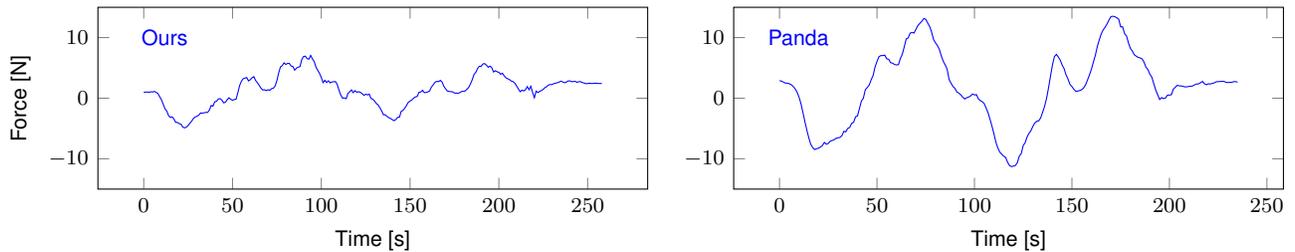

Different end-effectors (SenseGloves, Schunk SIH, Schunk SVH hand,
and corresponding 3D printed mounting adapters) are
mounted on each of the four involved force-torque sensors. In addition,
sensor bias results in barely usable raw sensor data.
Thus, each sensor is calibrated separately to compensate these effects.
20 data samples
from different sensor poses are collected. Each sample includes the
gravity vector in the sensor frame and the mean
of 100 sensor measurements from a static pose. A standard least squares
solver~\citep{4399184} is used to estimate the force-torque
sensor parameters, i.e. the force and torque bias and the mass and center of mass
of all attached components.
The same parameters including the additional mass and center of mass
transformation resulting by the force-torque
sensor itself is used to configure the built-in gravity compensation of
the Panda arms. The calibration is performed once after hardware changes
at the end-effectors or if the bias drift is too large.
This method does not compensate for
bias drift during usage but is sufficient for our application.

\section{Experiments}

We performed multiple experiments along with a small user study to evaluate the developed teleoperation system in our lab environment.

\subsection{Quantitative Experiments}

In a first experiment, we evaluated the operator arm workspace. 2959 different 6D
left hand poses were captured from a sitting person performing typical arm motions
with a VR tracker on their wrist.
In addition to hand poses with a fully extended arm, most of the poses are directly
in front of the person, likely to be performed during manipulation tasks.
First, the initial arm mounting pose (motivated by the avatar configuration) of the operator arm was evaluated. Each captured hand pose was marked as reachable if an inverse kinematic solution for the arm was found. In a second step, different arm mounting poses were sampled to find an optimal pose, maximizing the number of reachable hand poses (see \cref{fig:exp_arm_reach}). \cref{tab:arm_workspace} shows quantitative results.
The optimized arm mounting pose drastically increases the overlap between the human operator's and the avatar's arm workspace, but requires
a more complicated mounting setup.

\begin{table}
 \centering
 \caption{Operator arm workspace analysis}\label{tab:arm_workspace}
 \begin{tabular}{lrrr}
  \toprule
  Mounting Pose & Reached & Missed & Reached[\%]\\
  \midrule
  Initial   & 1795 & 1164 & 60.6\,\%\\
  Optimized & 2848 & 111 & 96.3\,\%\\
  \bottomrule
 \end{tabular}
\end{table}

In a second experiment, we evaluated the predictive limit avoidance module. Avatar arm position and velocity limits are  haptically displayed via joint forces to the operator. Since measured joint positions and velocities are afflicted with latency generated by network communication ($<$1\,ms) and motion execution using the Cartesian impedance controller, which can reach up to 200\,ms (see \cref{sec:avatar_arm_control}),
the operator control predicts the avatar arm joint configuration. \cref{fig:exp_model} shows the measured and predicted joint position of the first right arm joint. The prediction
compensates the delays, which allows for instantaneous feedback of the avatar arm limits to the operator.

In a third experiment, we investigated the forces and torques required to move the operator station arm, since this
directly affects operator fatigue.
We measured the forces and torques applied to the arm by reading the force-torque sensor measurements. The arm was
moved in a comparable manner once with only the Panda gravity compensation enabled (i.e. $\tau_{cmd} = 0$ see \cref{sec:operator_arm_control}) and a second time with our
arm force controller running. In \cref{fig:exp_ft}, the forces in the direction of one exemplary axis are shown.
The results show the advantage of using an external force-torque sensor to generate a more unencumbered feeling for the operator
while using the system.

\subsection{User Study}
Our goal was to create an immersive and intuitive feeling for operation at remote locations using our system.
Since humans have their very own preferences and subjective feelings of how good or intuitive certain control mechanisms
perform, we carried out a user study with untrained operators, comparing different telemanipulation approaches.
Due to the current COVID-19 pandemic, we were limited to immediate colleagues as subjects, which severely constrained
the scope of our study.

\begin{figure}
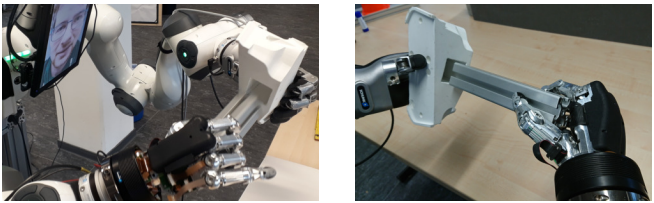

 \includegraphics[height=2.6cm]{images/experiments/task2.jpg}
 \hfill
 \includegraphics[height=2.6cm]{images/experiments/task3.jpg}
 \caption{User study with untrained operator: Both objects had to be grasped and the bar had to be inserted into the hole.}
 \label{fig:exp_task}
\end{figure}

A total of five participants were asked to perform a bimanual peg-in-hole manipulation task. First, two different objects had to be grasped: a small aluminum
bar and a 3D printed part with a hole. Afterwards, the bar should be inserted into the hole (see~\cref{fig:exp_task}). The task was challenging due to
very little friction between the finger and objects and tight tolerances which required precise insertion alignment.

Each participant performed the task three times with the following control modes:
\begin{enumerate}
 \item Operator station with haptic feedback enabled,
 \item Operator station with haptic feedback disabled, and
 \item VR controllers.
\end{enumerate}

Two HTV Vive VR controllers acted only as input devices: As long as the trigger
button was pressed, the corresponding avatar arm followed the controller movement.
A different button was programmed to toggle between a defined closed and open hand pose.
The tested control mode sequence was randomized for each participant. A maximum of 5\,min were granted to solve the task before it was marked as a failure. The participants were allowed to test each control mode about 1\,min before starting the measured test.

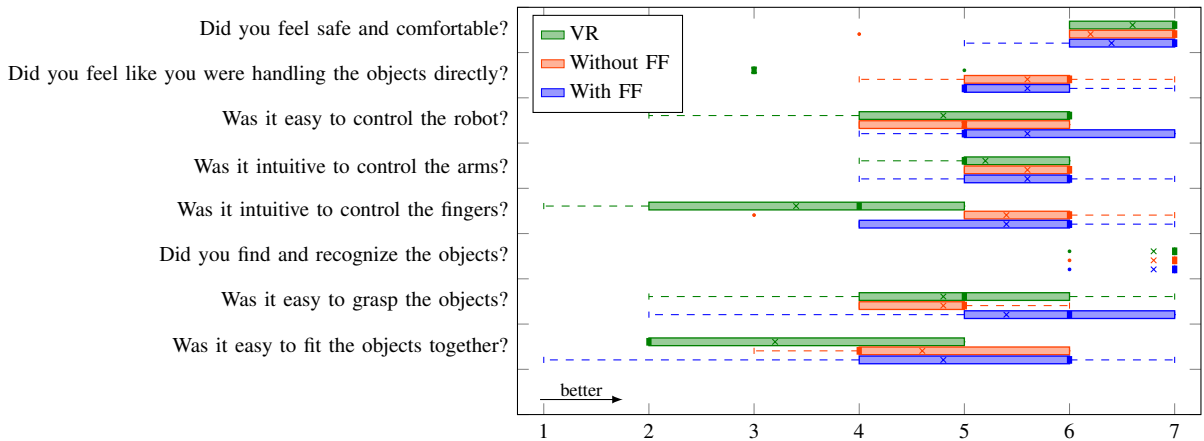
\begin{figure*}[h]
 ﻿\begin{tikzpicture}[font = \footnotesize, every mark/.append style={mark size=0.5pt}]
 \begin{axis}[
     name=plot,
     boxplot/draw direction=x,
     width=0.6\textwidth,
     height=7cm,
     boxplot={
         draw position={1.5 - 0.325/3 + 1.0*floor((\plotnumofactualtype + 0.001)/3) + 0.2*mod((\plotnumofactualtype + 0.001),3)},
         box extend=0.17,
         average=auto,
         every average/.style={/tikz/mark=x, mark size=1.5, mark options=black},
         every box/.style={draw, line width=0.5pt, fill=.!40!white},
         every median/.style={line width=2.0pt},
         every whisker/.style={dashed},
     },
     ymin=1,
     ymax=10,
     y dir=reverse,
     ytick={1,2,...,11},
     y tick label as interval,
     yticklabels={
Did you feel safe and comfortable?,
Did you feel like you were handling the objects directly?,
Was it easy to control the robot?,
Was it intuitive to control the arms?,
Was it intuitive to control the fingers?,
Did you find and recognize the objects?,
Was it easy to grasp the objects?,
Was it easy to fit the objects together?
     },
     y tick label style={
         align=center
     },
     xmin=0.75,
     xmax=7.25,
     xtick={1, 2 ,..., 7},
     xticklabels = {1, 2, ..., 7},
     cycle list={{green!50!black,orange!50!red,blue}},
     y dir=reverse,
     legend image code/.code={
         \draw [#1, fill=.!40!white] (0cm,-1.5pt) rectangle (0.3cm,1.5pt);
     },
     legend style={
         anchor=north west,
         at={($(0.0,1.0)+(0.2cm,-0.1cm)$)},
     },
     legend cell align={left},
 ]

 \addplot
 table[row sep=\\,y index=0] {
 data\\
 7\\7\\7\\6\\6\\
 };

 \addplot
 table[row sep=\\,y index=0] {
 data\\
 7\\7\\4\\6\\7\\
 };

 \addplot
 table[row sep=\\,y index=0] {
 data\\
 7\\7\\5\\6\\7\\
 };

 \addplot
 table[row sep=\\,y index=0] {
 data\\
 3\\1\\3\\3\\5\\
 };

 \addplot
 table[row sep=\\,y index=0] {
 data\\
 6\\4\\6\\5\\7\\
 };

 \addplot
 table[row sep=\\,y index=0] {
 data\\
 5\\6\\5\\5\\7\\
 };

 \addplot
 table[row sep=\\,y index=0] {
 data\\
 6\\2\\6\\4\\6\\
 };

 \addplot
 table[row sep=\\,y index=0] {
 data\\
 4\\5\\4\\6\\6\\
 };

 \addplot
 table[row sep=\\,y index=0] {
 data\\
 4\\7\\5\\5\\7\\
 };

 \addplot
 table[row sep=\\,y index=0] {
 data\\
 6\\4\\6\\5\\5\\
 };

 \addplot
 table[row sep=\\,y index=0] {
 data\\
 5\\5\\6\\6\\6\\
 };

 \addplot
 table[row sep=\\,y index=0] {
 data\\
 4\\7\\5\\6\\6\\
 };

 \addplot
 table[row sep=\\,y index=0] {
 data\\
 5\\2\\1\\5\\4\\
 };

 \addplot
 table[row sep=\\,y index=0] {
 data\\
 5\\7\\3\\6\\6\\
 };

 \addplot
 table[row sep=\\,y index=0] {
 data\\
 4\\7\\4\\6\\6\\
 };

 \addplot
 table[row sep=\\,y index=0] {
 data\\
 7\\7\\7\\6\\7\\
 };

 \addplot
 table[row sep=\\,y index=0] {
 data\\
 7\\7\\7\\6\\7\\
 };

 \addplot
 table[row sep=\\,y index=0] {
 data\\
 7\\7\\7\\6\\7\\
 };

 \addplot
 table[row sep=\\,y index=0] {
 data\\
 5\\4\\7\\2\\6\\
 };

 \addplot
 table[row sep=\\,y index=0] {
 data\\
 4\\5\\4\\5\\6\\
 };

 \addplot
 table[row sep=\\,y index=0] {
 data\\
 2\\7\\6\\5\\7\\
 };

 \addplot
 table[row sep=\\,y index=0] {
 data\\
 5\\2\\5\\2\\2\\
 };

 \addplot
 table[row sep=\\,y index=0] {
 data\\
 3\\4\\4\\6\\6\\
 };

 \addplot
 table[row sep=\\,y index=0] {
 data\\
 1\\7\\4\\6\\6\\
 };

 \legend{VR, Without FF, With FF}

 \end{axis}

 \draw[-latex] ($(plot.south west)+(0.3cm,0.2cm)$) -- ($(plot.south west)+(1.4cm,0.2cm)$)  node[midway,above,font=\scriptsize,inner sep=1pt] {better};

 \end{tikzpicture}
 \caption{Statistical results of our user questionnaire. We show the median, lower and upper quartile (includes interquartile range), lower and upper fence, outliers (marked with •) as well as the average value (marked with $\times$), for each aspect as recorded in our questionnaire.}
\vspace{-1em}
\label{fig:user_study}
\end{figure*}

\begin{table}
 \centering
 \caption{User study success rates and timings}\label{tab:success}
 \begin{tabular}{lrrr}
  \toprule
  Telemanipulation mode & Success & \multicolumn{2}{c}{Completion time [s]} \\
  \cmidrule (lr) {3-4}
              &         & \hspace{2em}Mean & StdDev \\
  \midrule
  1) Exoskeleton with feedback   & 4/5 & 119.0 & 117.1\\
  2) Exoskeleton w/o feedback   & 5/5 & 123.0 & 88.4\\
  3) VR controllers           & 3/5 & 126.3 & 25.8\\
  \bottomrule
 \end{tabular}
\end{table}

\cref{tab:success} shows the quantitative results of the user study. The time needed to successfully solve
the task is quite similar over the different telemanipulation modes. From these experiments, we
realized that the completion time was highly influenced by external factors, such as losing the object due to not enough
finger friction or different grasping and object handling solutions, which are unrelated to the used operator interface. In addition,
humans can easily compensate missing force feedback using visual feedback.
To generate a more meaningful statement based on performance scores, more test samples are needed.

In addition to these quantitative measurements, we asked the participants to
answer a short questionnaire about each telemanipulation mode with answers
from the 1-7 Likert scale (see \cref{fig:user_study}).
The results show that especially the feeling of handling the objects and
intuitive finger control was subjectively much better using
the Operator Station. Enabling the force and haptic feedback gives the highest
advantage when picking up the objects from the table.
This can be explained by the additional feedback indicating contact between
the hand and the table which cannot be perceived visually due to occlusions.
All participants reported to feel save and comfortable using the system.
Although the experiment time was limited, this suggests non-excessive cognitive load
on the operator.

Overall, the user study showed that our developed system is intuitive to use for
untrained operators. Even though the force and haptic feedback did not increase the success rate
of solving the task, it increases the immersive feeling as shown by the questionnaire.

\section{Discussion \& Conclusion}

This work presented a bimanual telemanipulation system consisting of an exoskeleton-based operator control station and an anthropomorphic avatar robot.
Both components communicate using our force and haptic feedback controller which allows safe and intuitive teleoperation
for both the operator and persons directly interacting with the avatar.
The control method is invariant to the kinematic parameters and uses only a common Cartesian hand frame for
commands and feedback. Using the predictive limit avoidance avatar model, arm limits for both the operator and avatar side
can be force displayed to the operator with low latency.
We evaluated the system using a user study on untrained operators
as well as lab experiments, which demonstrated the intuitiveness of our control method.

In future work, the workspace and haptic feedback will be further improved.

\renewcommand{\bibfont}{\normalfont\footnotesize}
\printbibliography

\end{document}